\documentclass[10pt,twocolumn,letterpaper]{article}

\usepackage{cvpr}              

\usepackage{graphicx}
\usepackage{amsmath}
\usepackage{amssymb}
\usepackage{booktabs}
\usepackage{epsfig}
\usepackage{subcaption}
\usepackage{multirow}
\usepackage{multicol}
\usepackage{times}
\usepackage{color}
\usepackage{arydshln}
\usepackage{xcolor}
\usepackage{colortbl}
\usepackage{array}

\usepackage[pagebackref,breaklinks,colorlinks]{hyperref}

\usepackage[capitalize]{cleveref}
\crefname{section}{Sec.}{Secs.}
\Crefname{section}{Section}{Sections}
\Crefname{table}{Table}{Tables}
\crefname{table}{Tab.}{Tabs.}
\definecolor{gbypink}{rgb}{0.99, 0.91, 0.95} 

\def\cvprPaperID{7439} 
\def\confName{CVPR}
\def\confYear{2023}

\begin{document}

\title{Privileged Prior Information Distillation for Image Matting}

\author{Cheng Lyu$^{1,2}$, Jiake Xie$^{3}$, Bo Xu$^{2,}$\thanks{Cheng Lyu, Jiake Xie, and Bo Xu contribute equally. This work was done when Cheng Lyu was interning at OPPO Research Institute. Bo Xu is the corresponding author.} , Cheng Lu$^{4}$, Han Huang$^{2}$,\\ Xin Huang$^{5}$, Ming Wu$^{1}$, Chuang Zhang$^{1}$, and Yong Tang$^{3}$\\
$^{1}$Beijing University of Posts and Telecommunications, $^{2}$OPPO Research Institute, \\$^{3}$PicUp.AI, $^{4}$Xmotors, $^{5}$University of Maryland, Baltimore County\\
}
\maketitle

\begin{abstract}
   Performance of trimap-free image matting methods is limited when trying to decouple the deterministic and undetermined regions, especially in the scenes where foregrounds are semantically ambiguous, chromaless, or high transmittance. In this paper, we propose a novel framework named Privileged Prior Information Distillation for Image Matting (PPID-IM) that can effectively transfer privileged prior environment-aware information to improve the performance of students in solving hard foregrounds. The prior information of trimap regulates only the teacher model during the training stage, while not being fed into the student network during actual inference. In order to achieve effective privileged cross-modality ($i.e.$ trimap and RGB) information distillation, we introduce a Cross-Level Semantic Distillation (CLSD) module that reinforces the trimap-free students with more knowledgeable semantic representations and environment-aware information. We also propose an Attention-Guided Local Distillation module that efficiently transfers privileged local attributes from the trimap-based teacher to trimap-free students for the guidance of local-region optimization. Extensive experiments demonstrate the effectiveness and superiority of our PPID framework on the task of image matting. In addition, our trimap-free IndexNet-PPID surpasses the other competing state-of-the-art methods by a large margin, especially in scenarios with chromaless, weak texture, or irregular objects.
\end{abstract}

\begin{figure}[t]
\centering
    \includegraphics[width=1.0\linewidth]{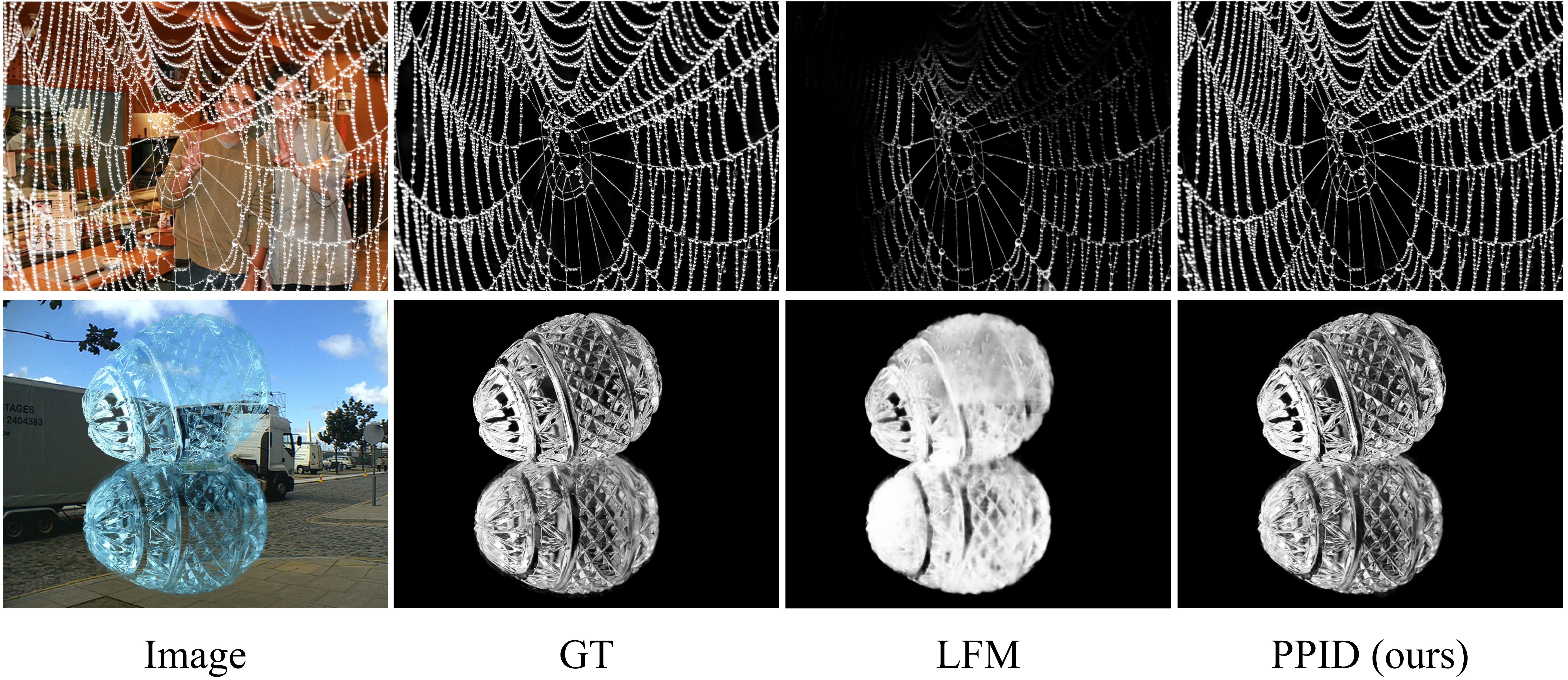}

\caption{Given challenging images of chromaless and weak texture objects, one SOTA trimap-free matting approach - LFM~\cite{zhang2019late}, fails to decouple the deterministic and undetermined regions. However, our PPID-based model is able to perform more accurate and fine-grained region decoupling for these hard foregrounds.}
\label{fig:intro}
\vspace{-10pt}
\end{figure}

\section{Introduction}
\label{sec:intro}
Image matting is one of the most fundamental computer vision tasks, which aims to separate the foreground objects from a single image or video stream. It has tremendous practical value for background replacement task in multimedia applications such as image/video entertainment creation, special-effect film-making and live video. To accurately estimate the opacity of each pixel inside foreground regions, the matting is generally formulated as an image/video frame composite problem, which solves the 7 unknown variables per pixel from only 3 known values:
\begin{equation}
    I_{i} = \alpha_{i} F_{i} + (1-\alpha_{i})B_{i},\ \alpha_{i} \in [0,1]
\label{E0}
\end{equation}
where $I_{i}$ refers to known 3-dimensional RGB color at pixel $i$, while foreground RGB color $F_{i}$, background RGB color $B_{i}$, and alpha matte estimation $\alpha_{i}$ are unknown. According to the above equation, $\alpha_{i}=1$ and $\alpha_{i}=0$ represent the deterministic foreground and background regions respectively, and $\alpha_{i}\in(0,1)$ refers to the undetermined regions with in-between pixels.  

To solve this highly challenging problem, the typical methods~\cite{xu2017deep,hou2019context,lu2019indices,li2020natural,liu2021tripartite} utilize trimap as a piece of environment-aware priori that marks the foreground, background and transition regions to locate the targets and reduce the solution spaces. Unfortunately, a high-quality trimap can not be obtained without tedious manual annotation effort and significant time costs, which limits its practical application in low-cost consumer products. Some trimap-free methods~\cite{zhang2019late,Qiao_2020_CVPR} are proposed to utilize the typical encoder-to-decoder structures that stem from segmentation and detection, $etc$, for alpha prediction without auxiliary cues. Although environmental saliency can be roughly predicted by borrowing such transitional network structure and its pre-trained feature, there still remain two tricky challenges in the trimap-free matting setting. First, previous trimap-free matting methods may fail to identify certain attributes when the foreground is semantically ambiguous, chromaless, or with high transmittance. For example in Row 2 of Figure~\ref{fig:intro}, due to the lack of strong texture and color hue, the previous trimap-free network~\cite{zhang2019late} fails to decouple the deterministic and undetermined regions of this highly transmissive ‘glass ball'. That is especially obvious in those upper pixels where the networks are confused by the background (sky) with similar colors. 

Second, for most of the existing trimap-free models, the learning of semantic mining often stems from upstream tasks such as segmentation. However, they may fail to identify the complete and meaningful foregrounds when such foregrounds are semantically ambiguous, rare, or spatially sparse,  \eg the cobweb in Row 1 of Figure~\ref{fig:intro}. Although several current methods try to construct pseudo-trimap or implicitly learn the transition region distribution as a decoupling effort by imposing local supervision on RGB features, they may fail to balance global and local matting quality (\eg texture similarity, location correlation, $etc.$) which leads to incomplete information mining.


To address the above challenges, we propose a novel Privileged Prior Information Distillation framework for Image Matting (PPID-IM). Our PPID-IM framework can effectively transfer privileged environment-aware prior information to improve the trimap-free models, leading to 1) more accurate region decoupling for hard foregrounds which are chromaless or with high transmittance 2) better balance between global and local matting quality. We rethink trimap-free matting via privileged information distillation, with the following motivations: a) Trimap provides critical environment-aware information which the trimap-free models may lack and 
focuses on the undetermined regions with weak texture and poor color hue. Consequently, it can be formulated as privileged modality in knowledge distillation framework. b) The privileged information (\eg features, extra modalities, $etc.$) distillation has been successfully applied in multiple tasks, \eg image classification~\cite{lopez2015unifying}, object detection~\cite{yang2022focal}, and image super-resolution~\cite{lee2020learning}, action detection~\cite{zhao2020privileged}, $etc$. Considering the listed successes, we borrow a similar idea to introduce trimap as privileged information for environment-aware knowledge distillation that can improve the performance of alpha matte prediction in a trimap-free setting, without modeling unreliable pseudo-trimap. 

To showcase this new matting framework, we employ multiple trimap-based matting models~\cite{xu2017deep,lu2019indices,li2020natural} as teachers to guide the training of their corresponding trimap-free students, which aims to demonstrate the generalization of our privileged prior information distillation for image matting. Each paired teacher and student share the same structure except for the inputs, where both RGB images and trimaps are given to the teacher and only RGB images for the student. To leverage effective privileged features for environment-aware information distillation, we introduce a Cross-Level Semantic Distillation (CLSD) module that guides each student layer to learn from more knowledgeable privileged features of the teacher model in addition to the corresponding layer distillation, for mining more effective environment-aware information. In addition, we propose an Attention-Guided Local Distillation module that efficiently transfers privileged local attributes from the trimap-based teacher to guide local region optimization for the trimap-free student. To justify our solutions, we conduct extensive experiments on multiple public datasets. The experimental results show that our proposed method surpasses all the state-of-the-art image matting approaches. Overall, the contributions of this paper are as follows:
\begin{itemize}
    \item We propose a novel Privileged Prior Information Distillation framework for Image Matting (PPID-IM) that can effectively transfer privileged  prior information to improve the trimap-free models, especially in scenes with chromaless, weak texture, or irregular objects.
    \item We introduce a Cross-Level Semantic Distillation (CLSD) module that complements the student networks with more sufficient environmental awareness and higher-level semantic feature representations.
    \item We also propose an Attention-Guided Local Distillation module to guide local region optimization of the trimap-free student by borrowing privileged local attributes from the trimap-based teacher.
    \item Extensive experiments demonstrate the effectiveness and superiority of our privileged prior information distillation method, outperforming the state-of-the-art (SOTA) trimap-free approaches on both synthetic and real-world images by a large margin. 
\end{itemize}
\begin{figure*}
    \centering
    \includegraphics[width=\linewidth]{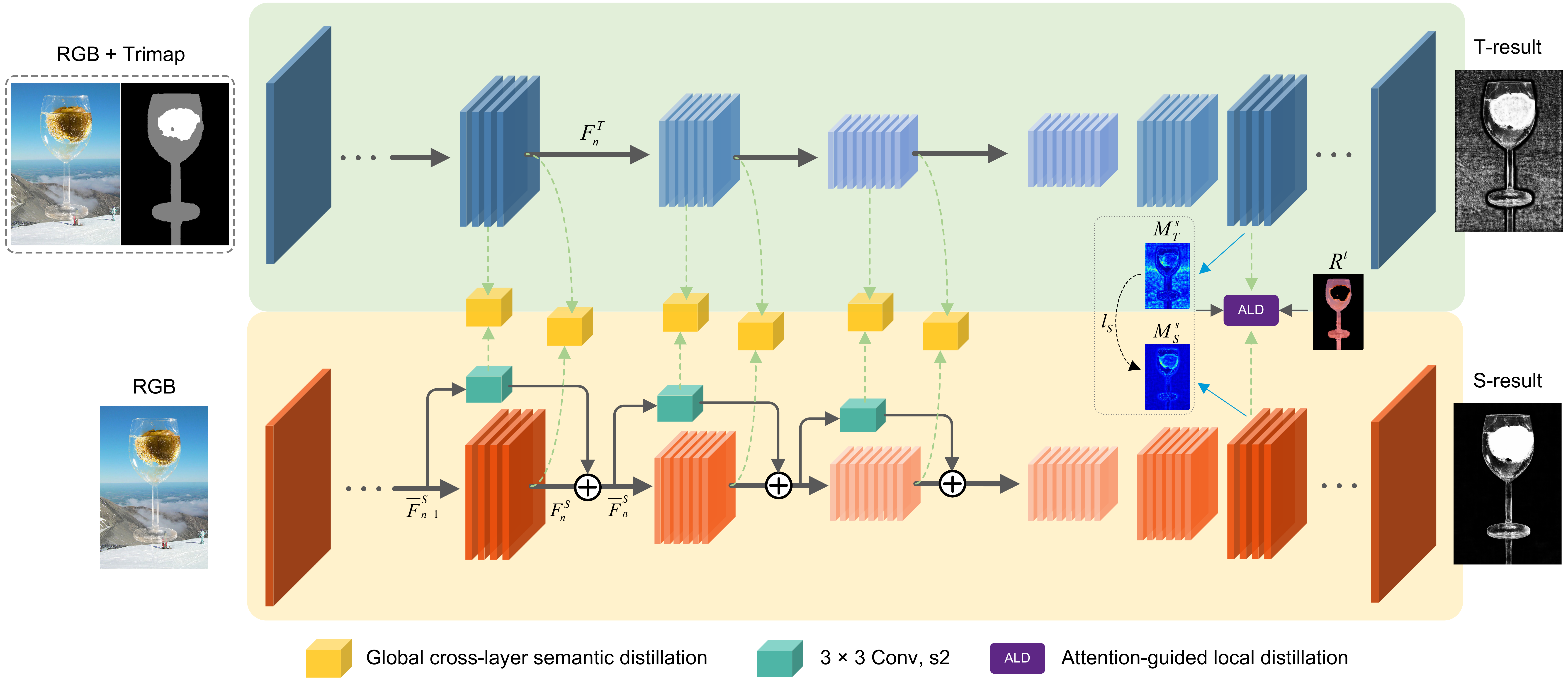}
    \caption{Architecture of the Privileged Prior Information Distillation for Image Matting (PPID-IM). The privileged information distillation across modalities (between trimap and RGB) is operated between the trimap-based teacher and its trimap-free student variant by an efficient Cross-Layer Semantic Distillation (CLSD) module. Subsequently, the privileged local attributes of the teacher network is further transferred to the student via an Attention-guided Local Distillation (ALD) module, for facilitating the student's local optimization.}
    \label{fig:architecture}
\end{figure*}

\section{Related works}
\subsection{Classic methods}
Classic foreground matting methods can be generally categorized into two approaches: sampling-based and propagation-based. Sampling-based methods~\cite{aksoy2017designing,chen2013image,shahrian2013improving,wang2007optimized,he2011global,chuang2001bayesian} sample the known foreground and background color pixels, and then extend these samples to achieve matting in other parts. Various sampling-based algorithms are proposed, $e.g.$ Bayesian matting~\cite{chuang2001bayesian}, optimized color sampling~\cite{wang2007optimized}, global sampling method~\cite{he2011global}, and comprehensive sampling~\cite{shahrian2013improving}. Propagation-based methods~\cite{levin2008spectral,chen2013knn,he2010fast,lee2011nonlocal,levin2007closed,sun2004poisson} reformulate the composite Eq.~\ref{E0} to propagate the alpha values from the known foreground and background into the unknown region, achieving more reliable matting results.~\cite{wang2008image} provides a very comprehensive review of various matting algorithms. 
\vspace{-5pt}
\subsection{Deep learning-based methods}
Classic matting methods are carefully designed to solve the composite equation and its variant versions. However, these methods heavily rely on chromatic cues, which leads to bad quality when the color of the foreground and background show small or no noticeable difference.

{\bf Trimap-based methods.} Initially, some attempts are made to combine deep learning networks with classic matting techniques, $e.g.$ closed-form matting~\cite{levin2007closed} and KNN matting~\cite{chen2013knn}. Cho \etal~\cite{cho2016natural} employ a deep neural network to improve the results of the closed-form matting and KNN matting. For facilitating the end-to-end training, Xu \etal~\cite{xu2017deep} propose a two-stage deep neural network (Deep Image Matting) based on SegNet~\cite{badrinarayanan2017segnet} for alpha matte estimation and contribute a large-scale composition image matting dataset (Adobe dataset) with ground truth foreground (alpha) matte. Lutz \etal~\cite{lutz2018alphagan} introduce a generative adversarial network (GAN) for natural image matting and improve the results of Deep Image Matting~\cite{xu2017deep}. Subsequently, a range of methods~\cite{hou2019context,lu2019indices,li2020natural,cai2019disentangled} have introduced different theoretical developments for matting performance improvements. Liu \etal~\cite{liu2021tripartite} propose a 3-branch network to mine and integrate complementary global information from the RGB image and trimap. 


{\bf Trimap-free methods.} Fully automatic matting without any auxiliary additional constraints (\eg user-annotated trimap) has also been studied. \cite{chen2018semantic} tries to predict the trimap first, followed by an alpha matting network. Zhang \etal~\cite{} propose a dual-decoder network for foreground and background classification, followed by a fusion branch to integrate the dual results. Qiao \etal~\cite{Qiao_2020_CVPR} employ spatial and channel-wise attention to integrating appearance cues and pyramidal features for trimap-free matting. Lin \etal~\cite{lin2022robust} propose a robust real-time matting method (RVM) training strategy that optimizes the network on both matting and segmentation tasks. Ke \etal~\cite{ke2020green} present a lightweight matting objective decomposition network (MODNet) and introduce an e-ASPP module for efficient multi-scale feature fusion. Besides, most current trimap-free methods may only mine limited environmental awareness information, due to a lack of privileged priori. This is the main reason why we propose the privileged prior information distillation for image matting.  

\subsection{Privileged information distillation methods}
Lopez-Paz \etal~\cite{lopez2015unifying} first combine Distillation~\cite{hinton2015distilling} and privileged information~\cite{vapnik2015learning} into generalized distillation, and extended it to unsupervised,
semi-supervised and multi-task learning scenarios. Li \etal~\cite{li2022cross} propose a cross-modality knowledge distillation model that leverages the additionally privileged depth to guide the training of the monocular visual odometry network. Wang \etal~\cite{wang2020pmd} propose a Privileged Modality Distillation Network that improves the RGB-based hand pose estimation by excavating the privileged information from depth prior during training. Zhao \etal~\cite{zhao2020privileged} consider future frames from the off-line teacher as privileged information to guide the online student for action detection, by knowledge distillation. According to the effectiveness in multiple tasks, we introduce privileged information distillation to improve the performance of trimap-free matting models.

\section{Methodology}
Our Privileged Prior Information Distillation for image matting (PPID-IM) is designed to effectively transfer the privileged prior information that can guide trimap-free students in capturing sufficient environmental awareness information and performing more accurate domain decoupling, especially in the scenarios with chromaless, weak texture or irregular objects.. During the training stage, the teacher network takes RGB images along with trimaps as inputs and the student network takes only RGB images as input. The overall architecture of the PPID-IM is shown in Figure~\ref{fig:architecture}, PPID consists of dual key components: cross-level semantic distillation (CLSD) and attention-guided local distillation (ALD) modules. 
\subsection{Cross-layer Semantic Distillation}
To effectively leverage privileged prior features for environment-aware information distillation, we propose a Cross-Level Semantic Distillation (CLSD) module that guides each student layer to learn higher-level privileged feature representations from the teacher in addition to the corresponding layer distillation. 

Inspired by multiple tasks~\cite{hu2018relation,hu2018squeeze,wang2018non,zhang2018context,cao2019gcnet} that need a global understanding of a visual scene, we believe that extracting the associations between pixels both within a region (\eg known, transition, or background regions) and across regions can lead to capturing more sufficient environment-aware information. We further employ the Global Context block (GCblock)~\cite{cao2019gcnet} to model the global association context of the teacher's feature representations and transfer them to both the corresponding layer and neighbor layer of the student network for effective privileged information distillation. The GCblock ($Gc(F)$)~\cite{cao2019gcnet} is formulated as follows:
\begin{equation}\label{E5}
\begin{aligned}
        Gc \big(F\big) = F+W_{v2} \big(ReLU \big(LN \big( W_{v1} \\ 
        \big(\sum_{j=1}^{N_p} \frac{{e^{W_k F_j}}}{ \sum_{m=1}^{N_p} e^{W_k F_M} } F_j \big) \big)\big)\big)
\end{aligned}
\end{equation}
where $W_k$, $W_{v1}$ and $W_{v2}$ denote linear transformation matrices, $LN$ denotes the layer normalization, $N_p$ denotes the number of positions in the feature map(\eg $N_p= H\cdot W$). The global feature transfer of CLSD can be formulated as:
\begin{equation}\label{E3}
\begin{aligned}
        L_{CLSD} &= \lambda_1 \cdot \sum\Big(Gc\big(F_n^T\big)-Gc\big(F_n^S\big)\Big)^2 \\
        &+\lambda_2 \cdot \sum\Big(Gc\big(F_n^T\big)-Gc\big(f^{3\times3}({\bar{F}}_{n-1}^S)\big)\Big)^2 \\
\end{aligned}
\end{equation}
\begin{equation}\label{E4}
\begin{aligned}
        \bar{F}_n^S  &= F_n^S+ f^{3\times3}(\bar{F}_{n-1}^S) \\
\end{aligned}
\end{equation}
where $Gc(F)$ represents the GCblock~\cite{cao2019gcnet}. $F_n^T$ and $F_n^S$ denote the feature maps of the teacher and student at the $n$-th layer, respectively. $f^{3\times3}$ is a $3 \times 3$ convolutional operation (stride=2). $\lambda_1$ and $\lambda_2$ are hyper-parameters for balancing the loss. We further update $F_n^S$ to $\bar{F}_n^S$ by merging it with the cross-layer distilled student feature $\bar{F}_{n-1}^S$ at $(n-1)$-th layer. 

\subsection{Attention-guided local distillation}

Most previous trimap-free methods are limited in terms of domain decoupling on local regions, due to the lack of prior constraints on the solution space. To address this situation, we propose an Attention-Guided Local Distillation (ALD) module that efficiently transfers privileged local attributes from the trimap-based teacher to guide the local region optimization for the trimap-free student. For efficient locally privileged information distillation, we generate a region mask to guide the transfer of effective pixel-level representations to the student, by the following formula: 
\begin{equation}\label{E8}
    R^t_{i,j}= \begin{cases}1,\quad if \left(i,j\right)\in t\\
    0,\quad otherwise\\
    \end{cases}
\end{equation}
where $t$ denotes the transition region, and $i$,$j$ refers to the pixel position in the feature map.

Specifically, we also introduce a spatial attention mask to further emphasize crucial information and suppress disturbing information about the target local regions. We follow CBAM~\cite{woo2018cbam} to respectively apply average-pooling ($AP$) and max-pooling ($MP$) along the channel and then feed the concatenated feature into a $7\times7$ convolution layer to generate the teacher's attention mask, as follows:
\begin{equation}\label{E7}
    M^{s}\left(F\right)= \sigma(f^{7\times7}\left(\frac{[AP\left(F\right);MP\left(F\right)]}{T}\right))
\end{equation}
where $\sigma$ denotes the Softmax function and $[ \cdot \ ;\cdot]$ is a concatenation operation along the channel axis. $T$ is the temperature hyper-parameter~\cite{hinton2015distilling} for distribution adjustment. Then the feature loss of ALD module can be formulated as follows:
\begin{equation}\label{E9}
\begin{aligned}
    L^{f}_{ALD}=\sum_{k=1}^C\sum_{i=1}^H\sum_{j=1}^W R^t_{ij}M_{i,j}^{s}\left(F_{k,i,j}^T-F_{k,i,j}^S\right)^2
\end{aligned}
\end{equation}
where $M^{s}$ denotes the teacher's attention map. $F^{T}$ and $F^{S}$ refer to the feature of the teacher and student, respectively. $C$, $H$, and $W$ denote the channel, height, and width of the given feature, respectively. Further, we introduce an attention distillation function $L^{a}_{ALD}$ that can guide the student to learn crucial information distribution of the transition regions from the teacher:
\begin{equation}\label{E10}
    L^{a}_{ALD}= l_s\left(R^{t}M_T^s, R^{t}M_S^s\right)
\end{equation}
where $M_T^s$ and $M_S^s$ denotes the attention maps of the teacher and student, respectively. $l_s$ is a L1 loss. Overall, the loss function of the ALD module ($L_{ALD}$) can be formulated by the combination of $L^{f}_{ALD}$ and $L^{a}_{ALD}$: 
\begin{equation}\label{E11}
    L_{ALD} = \alpha L^{f}_{ALD} + \beta L^{a}_{ALD}
\end{equation}
where $\alpha$ and $\beta$ are the hyper-parameters to balance the feature loss and attention loss. And then, the final loss function of the privileged prior information distillation is computed as:
\begin{equation}\label{E12}
    L_{distn}= L_{CLSD}+L_{ALD}
\end{equation}

\subsection{Loss function of Alpha Prediction}
According to Eq.~\ref{E0} and the definition of trimap, one image can be divided into three solution regions, $i.e.$ known foreground, background, and transition, where the scale of each region is commonly different. The large-scale regions tend to receive more attention in terms of loss, while the supervision in small-scale ones may be weakened due to their low pixel proportion, which will lead to an unbalance in alpha predictions of multiple regions. To address this challenge, we introduce a regional scaling mask to balance the supervision of each region, as follows:
\begin{equation}\label{E12}
    S_{ij}= \begin{cases}
    \frac{1}{N_f},\quad if\left(i,j\right)\in f\\
    \frac{1}{N_b},\quad if\left(i,j\right)\in b\\
    \frac{1}{N_t},\quad if\left(i,j\right)\in t\end{cases}
\end{equation}
where $i$,$j$ denotes the pixel position in the feature map, $N_f$, $N_b$ and $N_t$ represent the total number of pixels in the foreground, background and transition region, respectively. For the supervision of alpha prediction, we employ L1 loss ($L_{1}$), cross-entropy (CE) loss ($L_{ce}$), and gradient loss ($L_{grad}$) to reinforce different predictive attributes of the alpha matte, as follows:
\begin{equation}\label{E14}
\begin{aligned}
    L_{alpha}=\sum_{i=1}^{H_{\alpha}}\sum_{j=1}^{W_{\alpha}}  w_{ij}^{L1}S_{ij} L_{1}^i
    &+ \sum_{i=1}^{H_{\alpha}}\sum_{j=1}^{W_{\alpha}} w_{ij}^{ce} S_{ij} L_{ce}^i \\
    &+\sum_{i=1}^{H_{\alpha}}\sum_{j=1}^{W_{\alpha}} L_{grad}^i \\
\end{aligned}
\end{equation}
where $H_{\alpha}$ and $W_{\alpha}$ is the height and width of the final alpha prediction. $W_{ij}^{L1}$ and $W_{ij}^{ce}$ denote the weight of $L_{1}$ and $L_{ce}$ respectively and are set as follows:
\begin{equation}
    w_{ij}^{L1}= \begin{cases}1, \quad if \left(i,j\right) \in f\bigcup b \\
    2,\quad if \left(i,j\right) \in t
    \end{cases}
\end{equation}
\begin{equation}
    w_{ij}^{ce}= \begin{cases}1, \quad if \left(i,j\right) \in f\bigcup b \\
    0.5,\quad if \left(i,j\right) \in t
    \end{cases}
\end{equation}
where $f$, $b$, and $t$ denote the foreground, background and transition region, respectively. The $L_{1}$ loss is the absolute difference between the predicted and the ground truth alpha matte, and can recover both the global and local detail alpha values:
 \begin{equation}\label{E13}
    L_{1}^i = \left \|  \alpha_{i}-\alpha_{i}^{*} \right \|_{1}
\end{equation}
where $\alpha_{i}$ and $\alpha_{i}^{*}$ denote the student output and the ground truth alpha values at pixel $i$, respectively.

We follow~\cite{zhang2019late} to introduce the $L_{ce}$ loss to accelerate the convergence of the pixels in foreground and background regions towards to their targets, as follows: 
\begin{equation}
     L_{ce}^i= -[\alpha_{i}^{*} \cdot log\left(\alpha_{i}\right)+\left(1-\alpha_{i}^{*}\right)\cdot log\left(1-\alpha_{i}\right)]
\end{equation}

Similar to~\cite{zhang2019late}, we set a small weight for $L_{ce}$ and combine it with $L_{1}$ to supervise alpha prediction in the transition region. In addition, we utilize the gradient loss $L_{grad}$ to reduce the over-blurred alpha results, as follows:
\begin{equation}\label{E14}
    L_{grad}^i = \left|\nabla \alpha_{i} - \nabla \alpha_{i}^{*} \right|
\end{equation}
where $\nabla$ denotes the calculation of the gradient magnitude.

\subsection{Overall Training Loss}
Overall, we train the student network with the total loss as follows:
\begin{equation}
    L= L_{alpha}+ \gamma L_{distn}
\end{equation}
where $\gamma$ is the hyper-parameter to balance the alpha-prediction loss $L_{alpha}$ and the the privileged prior information distillation loss $L_{distn}$.

\begin{table*}[t]
\renewcommand\tabcolsep{3.8pt}
\begin{center}
\scalebox{0.96}{
\begin{tabular}{lc|c|cccc|cccc|cccc}
\toprule
\multicolumn{3}{c}{\bf Attribute}& \multicolumn{4}{c}{Transparent} & \multicolumn{4}{c}{Non-transparent} & \multicolumn{4}{c}{Whole Test set}\\
        \cmidrule(l){0-10} \cmidrule(l){12-15}
{\bf Methods}&{\bf Trimap}&{\bf Param}&SAD& MSE& Grad& Conn&SAD& MSE& Grad& Conn&SAD& MSE& Grad& Conn\\
\hline
DIM~\cite{xu2017deep}&\checkmark&25.58M&80.01&0.013&44.28&69.98&41.05&0.006&27.16&44.74&50.40&0.008&31.00&50.80\\
IndexNet~\cite{lu2019indices}&\checkmark&5.95M&61.73&0.007&27.18&50.13&39.96&0.004&25.38&36.45&45.80&0.005&25.90&43.70 \\
GCA~\cite{li2020natural}&\checkmark&25.13M&47.98&0.005&19.32&40.02&27.31&0.002&12.38&24.57&32.27&0.003&14.04&28.28\\
\midrule
LFM~\cite{zhang2019late}&-&37.91M&101.71&0.023&65.21&72.91&44.60&0.007&34.12&55.43&58.29&0.011&41.58&59.63\\
Hatt~\cite{Qiao_2020_CVPR}&-&34.26M&75.45&0.012&50.13&55.98&40.02&0.006&27.85&43.38&44.01&0.007&29.26&46.41\\
\midrule
DIM&-&25.58M&134.00&0.027&74.29&105.09&93.33&0.018&57.65&82.54&103.09&0.020&61.64&87.95\\
IndexNet&-&5.95M&122.08&0.023&67.48&74.11&57.08&0.008&36.70&58.79&72.68&0.012&44.08&62.47 \\
GCA&-&25.13M&105.99&0.022&62.30&70.41&64.53&0.010&37.07&53.87&74.48&0.013&43.13&57.84\\
\cdashline{1-15}[0.8pt/2pt]
\rowcolor{gbypink} DIM-PPID&-&29.12M&101.27&0.017&60.50&70.05&87.64&0.013&51.17&69.53&93.80&0.014&53.46&69.70\\
\rowcolor{gbypink} IndexNet-PPID&-&6.47M&\textbf{63.08}&\textbf{0.009}&\textbf{33.07}&\textbf{52.18}&\textbf{43.69}&\textbf{0.005}&\textbf{25.59}&\textbf{38.39}&\textbf{48.35}&\textbf{0.006}&\textbf{27.39}&\textbf{41.70} \\
\rowcolor{gbypink} GCA-PPID&-&26.60M&78.11&0.013&46.82&68.25&50.70&0.007&32.93&49.28&57.27&0.009&36.90&53.83\\
\bottomrule
\end{tabular}
}
\end{center}
\caption{The quantitative results on Composition-1K~\cite{xu2017deep}.}
\label{tab:aim}
\vspace{-6pt}
\end{table*}

\begin{table*}[t]
\renewcommand\tabcolsep{3.8pt}
\begin{center}
\scalebox{0.96}{
\begin{tabular}{lc|c|cccc|cccc|cccc}
\toprule
\multicolumn{3}{c}{\bf Attribute}& \multicolumn{4}{c}{Transparent} & \multicolumn{4}{c}{Non-transparent} & \multicolumn{4}{c}{Whole Test set}\\
        \cmidrule(l){0-10} \cmidrule(l){12-15}
{\bf Methods}&{\bf Trimap}&{\bf Param}&SAD& MSE& Grad& Conn&SAD& MSE& Grad& Conn&SAD& MSE& Grad& Conn\\
\hline
DIM~\cite{xu2017deep}&\checkmark&25.58M&89.41&0.019&51.25&74.03&23.19&0.004&20.32&19.81&36.44&0.007&26.51&30.65\\
IndexNet~\cite{lu2019indices}&\checkmark&5.95M&97.61&0.017&41.27&74.44&18.79&0.003&16.62&15.96&34.55&0.006&21.55&27.66 \\
GCA~\cite{li2020natural}&\checkmark&25.13M&97.24&0.020&56.61&76.03&23.72&0.005&21.14&20.02&38.42&0.008&28.23&31.22\\
\midrule
LFM~\cite{zhang2019late}&-&37.91M&113.53&0.022&83.10&105.76&59.72&0.011&52.61&47.93&69.74&0.013&58.85&72.11\\
Hatt~\cite{Qiao_2020_CVPR}&-&34.26M&104.57&0.019&69.72&87.33&35.08&0.007&34.53&40.58&48.98&0.009&41.57&49.93\\
\midrule
DIM&-&25.58M&138.42&0.031&88.16&128.53&63.75&0.012&53.71&62.73&78.68&0.016&60.60&75.89\\
IndexNet&-&5.95M&134.99&0.030&86.47&125.38&57.29&0.010&52.42&59.33&72.83&0.014&59.03&72.54 \\
GCA&-&25.13M&113.81&0.028&84.38&101.78&46.63&0.009&37.44&41.43&60.07&0.013&46.83&53.50\\
\cdashline{1-15}[0.8pt/2pt]
\rowcolor{gbypink} DIM-PPID&-&29.12M&105.38&0.020&71.64&83.81&30.49&0.005&27.11&25.82&45.47&0.008&36.01&37.42\\
\rowcolor{gbypink} IndexNet-PPID&-&6.47M&\textbf{100.11}&\textbf{0.018}&\textbf{63.59}&\textbf{72.31}&\textbf{22.52}&\textbf{0.004}&\textbf{18.35}&\textbf{18.66}&\textbf{37.35}&\textbf{0.006}&\textbf{27.40}&\textbf{29.40} \\
\rowcolor{gbypink} GCA-PPID&-&26.60M&102.73&0.019&70.05&85.51&30.33&0.005&25.60&23.51&44.81&0.008&34.49&35.91\\
\bottomrule
\end{tabular}}
\end{center}
   
\caption{The quantitative results on Distinction-646~\cite{Qiao_2020_CVPR}.}
\label{tab:dist}
\end{table*}1

\section{Experiments}
\subsection{Datasets}
We first describe the datasets used for training and testing. Subsequently, we compare our results based on privileged information distillation with the trimap-based teacher models and existing state-of-the-art (SOTA) trimap-free image matting algorithms. Finally, we conduct ablation experiments to show the effectiveness of each component.

\textbf{Adobe image matting (AIM)~\cite{xu2017deep}.} The training set consists of 431 foreground objects and each of them is composited over 100 random COCO~\cite{lin2014microsoft} images to produce 43.1k composited training images. For the test set, we first composite each foreground from the test set with 20 random VOC~\cite{everingham2010pascal} images to produce 1k composited testing images (Composition-1K). Then we split Composition-1K into two groups (240 and 760 images, respectively) based on the critical attributes of our interest that separate levels of difficulty, called transparent and non-transparent.

\textbf{Distinction-646~\cite{Qiao_2020_CVPR}.} It includes 596 and 50 foreground objects in training and test sets, respectively. We enforce the same rule, composited ratio, and grouping style with the AIM datasets that split the test set into two groups consists of 200 and 800 images, respectively. 

\subsection{Implementation Details}
We implement the privileged prior information distillation on different matting models, including DIM~\cite{xu2017deep}, IndexNet~\cite{lu2019indices}, and GCA~\cite{li2020natural} to evaluate the general applicability of our method. The teacher and the student use the same models but differ only in the inputs (RGB images and trimaps for the teacher, only RGB images for the student). All the experiments are conducted with Pytorch\cite{paszke2019pytorch}. We employ the SGD optimizer with the momentum (0.9) and the weight decay (0.0005) for 50k iterations. The learning rate is initialized to be 0.01 and is multiplied by $\left(1-\frac{iter}{max-iter}\right)^{0.9}$. More implementation details are provided in supplementary material.


\begin{table*}[t]
\renewcommand\tabcolsep{3.2pt}
\begin{center}
\scalebox{0.92}{
\begin{tabular}{c|cc|cccc|cccc|cccc}
\toprule
\multicolumn{3}{c}{\bf Attribute}& \multicolumn{4}{c}{Transparent} & \multicolumn{4}{c}{Non-transparent}& \multicolumn{4}{c}{Whole Test set}\\
\midrule
{\bf Baselines}&$SD$&$CLSD$&SAD&MSE&Grad&Conn&SAD&MSE&Grad&Conn&SAD&MSE&Grad&Conn\\
\hline
\multirow{3}{*}{DIM~\cite{xu2017deep} + $ALD$}&&&116.23&0.024&70.11&85.73&90.82&0.017&55.32&71.93&96.92&0.019&58.87&75.24\\
&\checkmark&&103.28&0.020&66.82&70.29&89.24&0.015&54.81&70.46&92.61&0.016&57.69&70.42\\
\rowcolor{gbypink}&&\checkmark&101.27&0.017&60.50&70.05&87.64&0.013&51.17&69.53&93.80&0.014&53.46&69.70\\
\midrule
\multirow{3}{*}{IndexNet~\cite{lu2019indices} + $ALD$}&&&104.34&0.021&67.52&72.58&55.37&0.008&29.05&56.44&67.12&0.012&38.28&60.31\\
&\checkmark&&80.62&0.016&58.83&70.01&52.66&0.007&26.90&52.49&59.37&0.009&34.56&56.69\\
\rowcolor{gbypink}&&\checkmark&\textbf{63.08}&\textbf{0.009}&\textbf{33.07}&\textbf{52.18}&\textbf{43.69}&\textbf{0.005}&\textbf{25.59}&\textbf{38.39}&\textbf{48.35}&\textbf{0.006}&\textbf{27.39}&\textbf{41.70}\\
\midrule
\multirow{3}{*}{GCA~\cite{li2020natural} + $ALD$}&&&103.72&0.020&63.52&69.06&62.01&0.010&36.85&49.89&72.02&0.012&43.25&54.49\\
&\checkmark&&84.26&0.018&61.66&68.87&55.81&0.009&34.53&49.71&62.64&0.011&41.04&54.31\\
\rowcolor{gbypink}&&\checkmark&78.11&0.013&46.82&68.25&50.70&0.007&32.93&49.28&57.27&0.009&36.90&53.83\\
\bottomrule
\end{tabular}}
\end{center}
\caption{Ablation study of the cross-layer semantic distillation (CLSD) module on AIM~\cite{xu2017deep}. {\bf Com:} components of our PPID-IM; {\bf M}: metrics; {\bf BL:} baselines; $SD:$ distillation is only performed between the same layers; $CLSD:$ cross-layer semantic distillation.}
\label{tab:ablation-1}
\end{table*}

\begin{table}[t]
\renewcommand\tabcolsep{3.2pt}
\begin{center}
\scalebox{0.93}{
\begin{tabular}{c|cc|cccc}
\toprule
{\bf Baselines}&$LD$&$ALD$&SAD&MSE&Grad&Conn\\
\hline
\multirow{3}{*}{DIM~\cite{xu2017deep}}&&&120.08&0.027&80.76&109.25\\
&\checkmark&&60.19&0.020&45.56&66.24\\
\rowcolor{gbypink}&&\checkmark&51.68&0.015&30.40&51.75\\
\midrule
\multirow{3}{*}{IndexNet~\cite{lu2019indices}}&&&70.59&0.025&69.88&70.56\\
&\checkmark&&58.72&0.021&43.65&61.95\\
\rowcolor{gbypink}&&\checkmark&55.39&0.017&39.02&56.11\\
\midrule
\multirow{3}{*}{GCA~\cite{li2020natural}}&&&114.37&0.027&78.14&108.98\\
&\checkmark&&56.02&0.018&37.25&57.40\\
\rowcolor{gbypink}&&\checkmark&51.67&0.014&28.62&49.28\\
\bottomrule
\end{tabular}
}
\end{center}
\caption{Ablation study of the attention-guided local distillation (ALD) module on AIM~\cite{xu2017deep}. $LD:$ local distillation w/o attention guidance; $ALD:$ attention-guided local distillation.}
\label{tab:ablation-2}
\end{table}

\subsection{Comparative Study}
We conduct a comparative study on two public composition benchmarks: Adobe Image Matting (AIM)~\cite{xu2017deep} and Distinction-646~\cite{Qiao_2020_CVPR}. We report mean square error (MSE), the sum of the absolute difference (SAD), spatial-gradient (Grad), and connectivity (Conn) between predicted and ground truth alpha mattes. Lower values of these metrics indicate better-estimated alpha matte. To fairly compare, the metrics are computed on the entire image. 

We summarize the performance comparison in above two groups between our trimap-free students, the trimap-based teachers ($i.e.$ DIM~\cite{xu2017deep}, IndexNet~\cite{lu2019indices}, GCA~\cite{li2020natural}),  their trimap-free baselines (w/o trimap, and end-to-end training), and the state-of-the-art trimap-free models that include LFM~\cite{zhang2019late} and Hatt~\cite{Qiao_2020_CVPR}. For methods without publicly available codes, we follow
their papers to reproduce the results with due diligence.

\begin{figure}
    \centering
    \includegraphics[width=\linewidth]{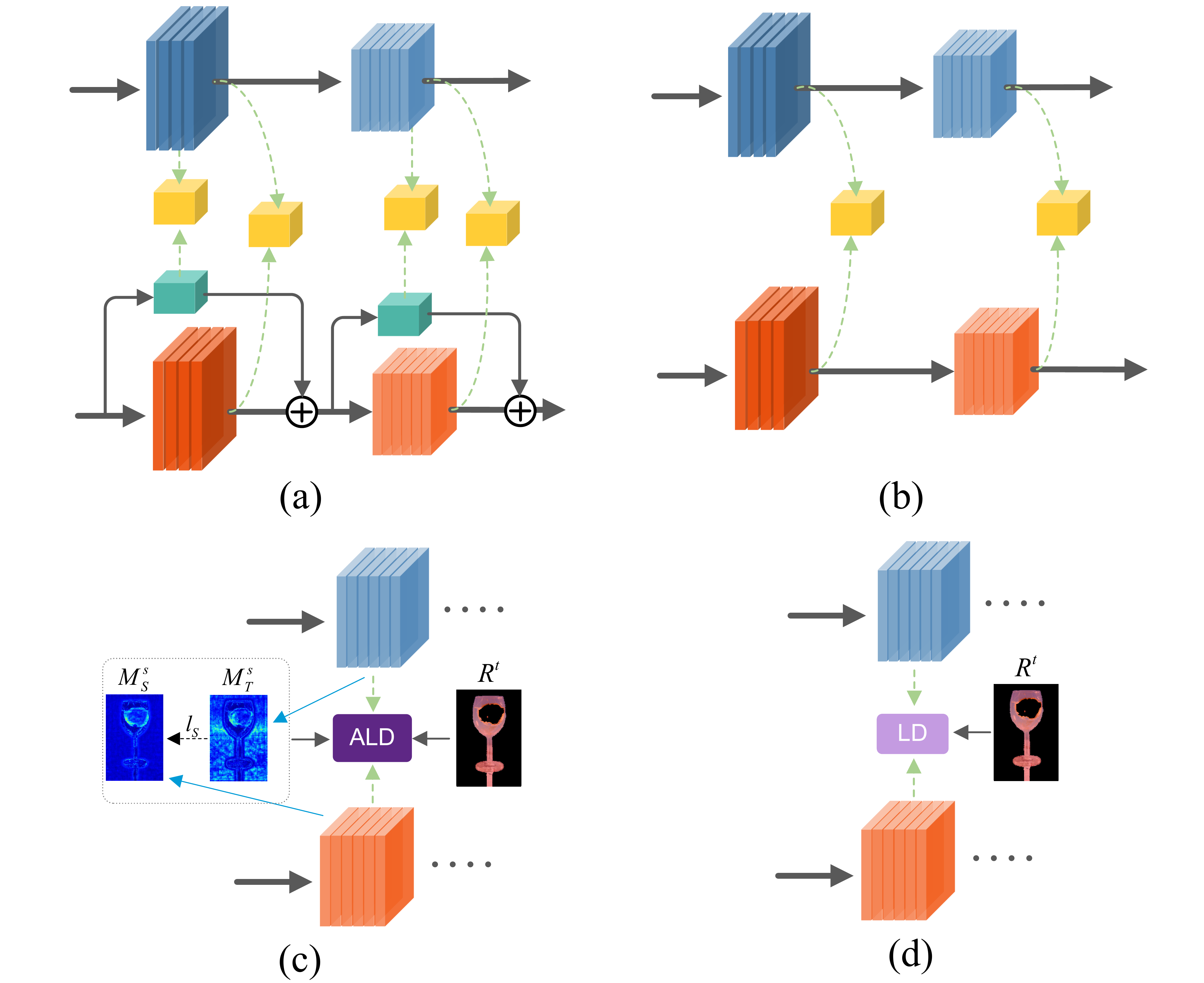}
    \caption{Ablation settings for PPID: (a) $CLSD$; (b) $SD$; (c) $ALD$; (d) $LD$.}
    \label{fig:ablation}
\end{figure}

Table~\ref{tab:aim} and~\ref{tab:dist} show the quantitative results of our PPID-based models with other SOTA trimap-free methods on Composition-1k~\cite{xu2017deep} and Distinction-646~\cite{Qiao_2020_CVPR}. We notice that our privileged prior information distillation framework significantly improves the performance of trimap-free baseline models in region decoupling, especially on the transparent attribute groups of both datasets. Compared to the existing SOTA methods, more light-weighted trimap-free baseline models can achieve more significant performance based on effective environment-aware information complements and guided local attribute optimization from our PPID framework.  Particularly, our trimap-free IndexNet-PPID surpasses the other competing methods~\cite{xu2017deep,Qiao_2020_CVPR} by a large margin, both on transparent and non-transparent attributes. 
\begin{figure*}[t]
  \centering
  \includegraphics[width=0.96\linewidth]{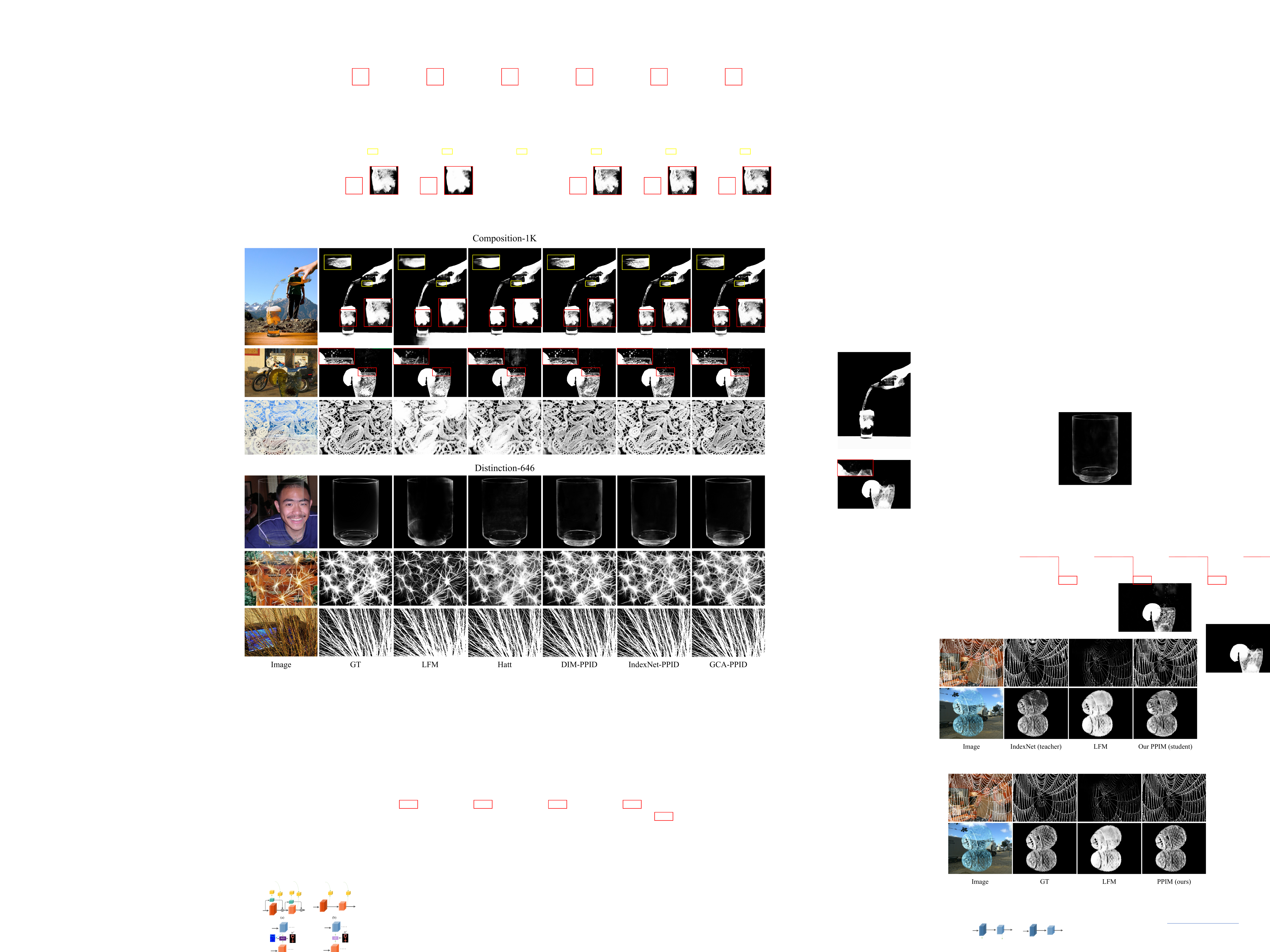}
  \caption{Visual comparison on public composition datasets. Trimap-free methods: LFM~\cite{zhang2019late}, Hatt~\cite{Qiao_2020_CVPR}, our DIM-PPID, IndexNet-PPID, and GCA-PPID.}\label{fig:visual_1}
  
\end{figure*}
\vspace{6pt}
\subsection{Ablation Study}
To validate the effectiveness of our key components in privileged information distillation, we conduct ablation study under the following settings: (a) $CLSD$: cross-layer semantic distillation (CLSD); (b) $SD$: semantic distillation is only performed between the same layers; (c) $ALD$: attention-guided local distillation; (d) $LD$: local distillation w/o attention guidance. The module settings of the ablation study are also illustrated in Figure~\ref{fig:ablation}. 

\textbf{CLSD $vs.$ SD.} We report the quantitative comparison results of our privileged information distillation on both transparent and non-transparent groups of the AIM~\cite{xu2017deep} dataset, with and without the privileged semantic distillation components (CLSD or SD). As summarized in Table~\ref{tab:ablation-1}, applying either CLSD or SD can significantly contribute to environmental awareness enhancement of the trimap-free baseline networks ($i.e.$ DIM~\cite{xu2017deep}, IndexNet~\cite{lu2019indices}, and GCA~\cite{li2020natural}), and the alpha prediction performance is further improved on both transparent and non-transparent groups by introducing the CLSD module. That is because our proposed cross-layer semantic distillation mechanism can complement more sufficient environment-aware information by guiding each student layer to additionally mine higher-level semantic context.

\textbf{ALD $vs.$ LD.} We perform quantitative comparisons on our PPID-based models~\cite{xu2017deep,lu2019indices,li2020natural} with or without the ALD module under three settings, $i.e.$ w/o $ALD$, with $LD$, and with $ALD$. To evaluate the effectiveness of the ALD module on local attribute optimization, we compute the metrics on the transition region of each image in AIM~\cite{xu2017deep}. As shown in Table~\ref{tab:ablation-2}, the proposed local distillation modules ($ALD$ and $LD$) improve the performance in local detail predictions, it offers further gain after introducing the spatial attention guidance that forces the students to focus on the crucial pixels. Additionally, we demonstrate the significant performance gain after combining CLSD and ALD. Some representative visualizations are provided in Figure~\ref{fig:visual_1}, which also illustrate the effectiveness of our PPID-IM, especially in the scenarios with foregrounds that are semantically ambiguous (Row 3, 5, and 6), chromaless (Row 1, 2, and 4), or irregular (Row 3, 5, and 6).   
\section{Conclusion}
\vspace{-6pt}
In this paper, we propose a privileged prior information distillation framework for image matting (PPID-IM), that aims to effectively transfer privileged prior information from the trimap-based teachers to their poor environment-aware trimap-free student models. We also introduce a Cross-Level Semantic Distillation (CLSD) module that complements the student networks with both environmental awareness and higher-level semantic feature representations, for facilitating the cross-modality information distillation. Further, an Attention-Guided Local Distillation (ALD) module is proposed to guide local region optimization for the students by efficiently transferring privileged local attributes and crucial information distribution from the teachers. Extensive experiments demonstrate the effectiveness and superiority of our PPID on image matting.


\appendix


We provide additional details in this supplementary material. In Sec. \ref{sec:impl}, we describe the details of our implementation. 
In Sec. \ref{att_map}, we compare the spatial attention maps results.
In Sec. \ref{alpha}, we provide more comparative visualizations on Composition-1K~\cite{xu2017deep} and Distinction-646~\cite{Qiao_2020_CVPR} datasets. In addition, we also illustrate visualization results on real-world images to demonstrate the generalizability of our PPID framework.
\vspace{-6pt}
\section{Implementation}
\label{sec:impl}
\vspace{-6pt}
For training, all input images are randomly cropped to $512\times512$, $640\times
640$, and $960\times960$. Then, they are resized to a resolution of $512\times512$. And we follow the same random flipping and random trimap dilation strategies as DIM\cite{xu2017deep}. All training samples are created on-the-fly. We provide each attribute group (transparent or non-transparent) list of test sets in ‘testset$\_$trn$\_$non-trn$\_$split.txt’ file.

We use 4 Tesla V100 GPUs for training multiple paired teacher-student models based on our PPID-IM framework, including DIM\cite{xu2017deep}, IndexNet\cite{lu2019indices}, and GCA\cite{li2020natural} that employ ImageNet\cite{deng2009imagenet} pre-trained VGG16\cite{simonyan2014very}, ImageNet pre-trained MobileNetV2\cite{sandler2018mobilenetv2}, and ImageNet pre-trained ResNet\cite{he2016deep} as backbones, respectively. The batch size is set to 64. The loss weight settings of different models on the AIM~\cite{xu2017deep} and Distinctions-646~\cite{Qiao_2020_CVPR} are shown in Table~\ref{tab:sub-1}.


\begin{table}[t]
\begin{center}
\scalebox{0.92}{
\begin{tabular}{l|ccccc}
\toprule
\multicolumn{1}{c}{\bf Attribute}& \multicolumn{5}{c} {Loss weight hyper-parameters} \\
 \cmidrule(l){0-1} \cmidrule(l){1-6}
 
{models}&$\lambda_1$&$\lambda_2$&$\alpha$&$\beta$&$\gamma$\\
\midrule 
 DIM&0.2&1.0&0.02&0.002&1.0\\
 \midrule
 IndexNet&0.2&1.0&0.002&0.0002&1.0\\
 \midrule
 GCA&0.2&1.0&0.002&0.0002&1.0\\
\bottomrule
\end{tabular}}
\end{center}
\caption{Loss weights settings.}
\label{tab:sub-1}
\vspace{-10pt}
\end{table}

\begin{table}[t]
\renewcommand\tabcolsep{3.2pt}
\begin{center}
\scalebox{0.92}{
\begin{tabular}{c|cc|cccc}
\toprule
{\bf Baselines}&$L^{f}_{ALD}$&$L^{a}_{ALD}$&SAD&MSE&Grad&Conn\\
\hline
\multirow{3}{*}{DIM~\cite{xu2017deep}}&-&-&120.08&0.027&80.76&109.25\\
&\checkmark&-&57.22&0.017&38.23&59.61\\
&-&\checkmark&78.39&0.019&54.81&72.05\\
\rowcolor{gbypink}&\checkmark&\checkmark&51.68&0.015&30.40&51.75\\
\midrule
\multirow{3}{*}{IndexNet~\cite{lu2019indices}}&-&-&70.59&0.025&69.88&70.56\\
&\checkmark&-&60.40&0.019&46.81&61.01\\
&-&\checkmark&66.87&0.023&58.43&65.29\\
\rowcolor{gbypink}&\checkmark&\checkmark&55.39&0.017&39.02&56.11\\
\midrule
\multirow{3}{*}{GCA~\cite{li2020natural}}&-&-&114.37&0.027&78.14&108.98\\
&\checkmark&-&58.06&0.017&41.77&60.90\\
&-&\checkmark&66.15&0.022&56.25&63.57\\
\rowcolor{gbypink}&\checkmark&\checkmark&51.67&0.014&28.62&49.28\\
\bottomrule
\end{tabular}}
\end{center}
\caption{Analysis of $L^{f}_{ALD}$ and $L^{a}_{ALD}$ in the attention-guided local distillation (ALD) module on AIM~\cite{xu2017deep}.}
\label{tab:ald}
\vspace{-10pt}
\end{table}

\begin{figure*}[t]
  \centering
  \includegraphics[width=0.9\linewidth]{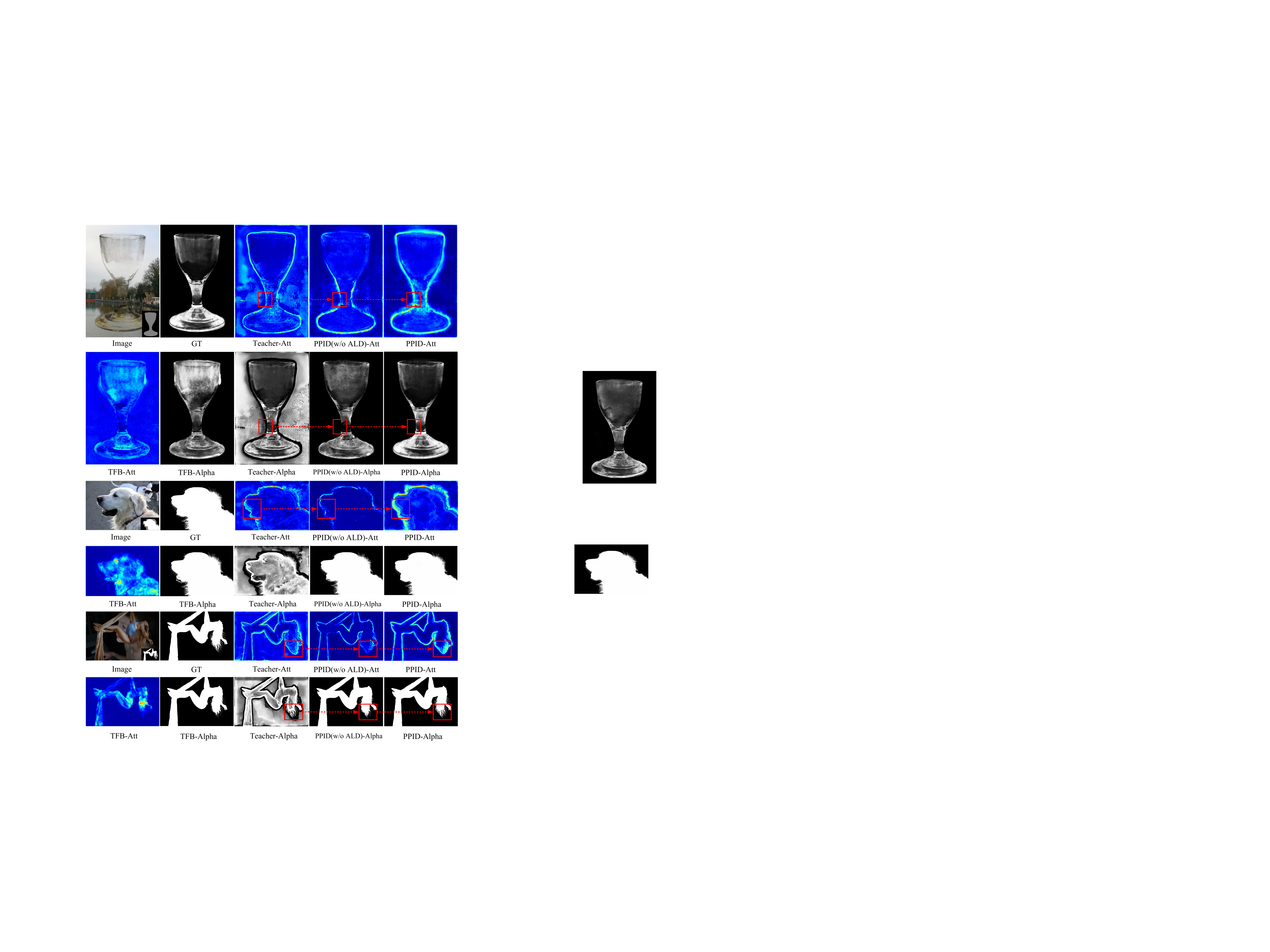}
  \caption{Representative visualizations on the effectiveness of the ALD module. Teacher: IndexNet. TFB denotes trimap-free baseline.}\label{fig:visual_att}
\end{figure*}

\section{Analysis of components in ALD}
\label{att_map}
$L^{f}_{ALD} \& L^{a}_{ALD}$. We report the quantitative comparison results of the Attention-Guided Local Distillation (ALD) module under either the feature loss $L^{f}_{ALD}$ or the attention distillation loss $L^{a}_{ALD}$. We compute the metrics on the transition region of each image in AIM~\cite{xu2017deep}. As shown in Table~\ref{tab:ald}, both attention-guided feature distillation ($L^{f}_{ALD}$) and attention map distillation ($L^{a}_{ALD}$) can improve the performance of our PPID on local optimization. After the combination of $L^{f}_{ALD}$ and $L^{a}_{ALD}$, the performance gains further, which demonstrates the effectiveness and necessity of the proposed components in the ALD module. We also provides representative visualizations in Figure~\ref{fig:visual_att} to further illustrate that the trimap-free students can learns the distribution of more crucial local information from the teachers through the attention-guided local distillation.


\begin{figure*}[t]
  \centering
  \includegraphics[width=1.0\linewidth]{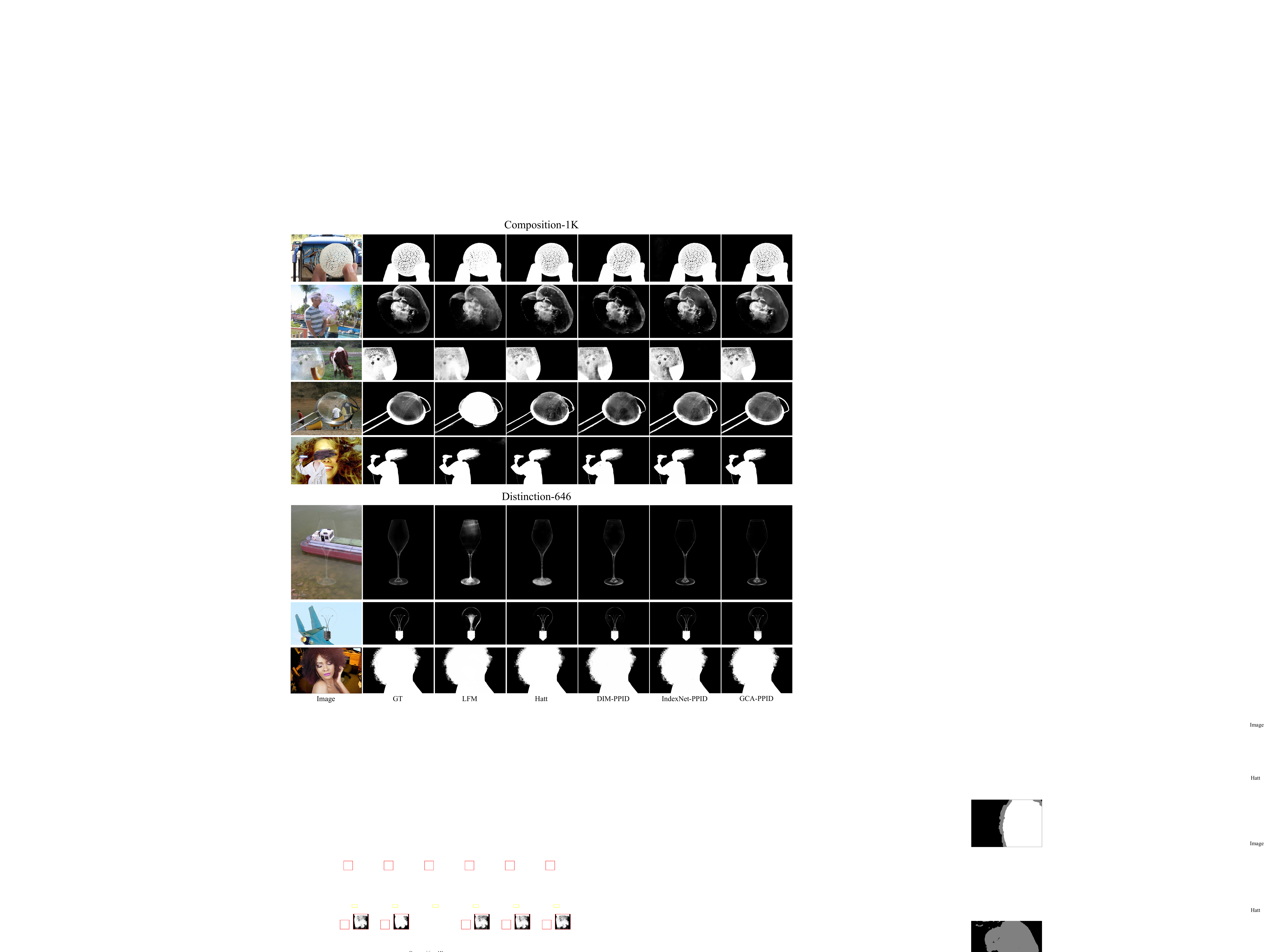}
  \caption{More Visual comparisons on Composition-1K~\cite{xu2017deep} and Distinction-646~\cite{Qiao_2020_CVPR}.}\label{fig:visual_1}
\end{figure*}

\begin{figure*}[t]
  \centering
  \includegraphics[width=1.0\linewidth]{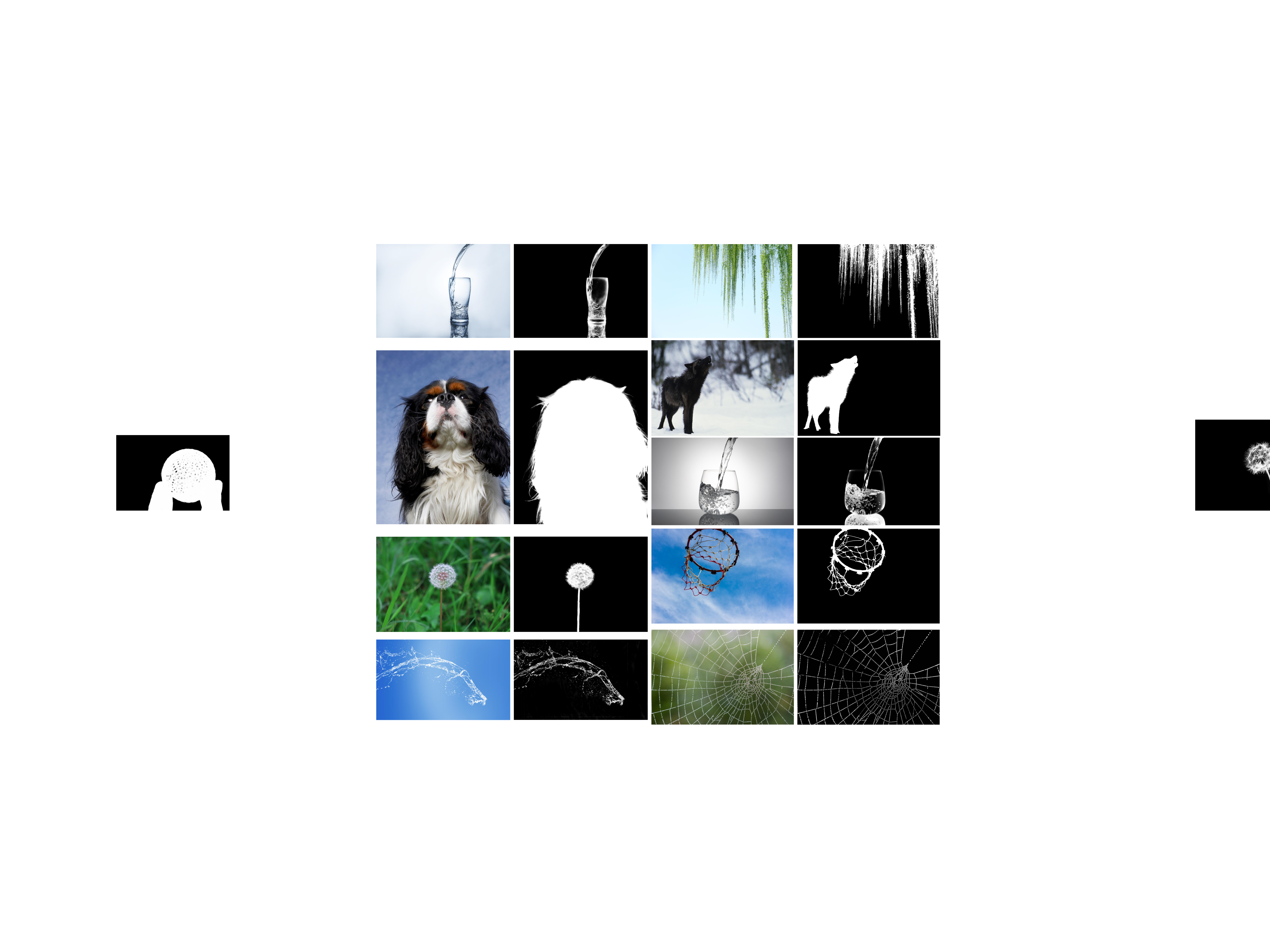}
  \caption{The alpha mattes produced by our PPID framework on real images.}\label{fig:visual_real}
\end{figure*}
\vspace{-6pt}
\section{More Visualization Results}
\label{alpha}
\vspace{-6pt}
We display more representative visualizations on Composition-1K~\cite{xu2017deep}, Distinction-646~\cite{Qiao_2020_CVPR}, and real-world images. Performance comparisons in
Figure~\ref{fig:visual_1} and alpha mattes produced by our PPID in Figure~\ref{fig:visual_real} demonstrate the effectiveness and generalization of our
privileged prior information distillation framework for image matting (PPID-IM), especially in scenarios with
chromaless, weak texture, or irregular objects.

\end{document}